%% file: main.tex
\documentclass[11pt]{article}

\usepackage[final]{acl}

\usepackage{times}
\usepackage{latexsym}
\usepackage{amssymb}
\usepackage{enumerate}

\usepackage[T1]{fontenc}

\usepackage[utf8]{inputenc}

\usepackage{microtype}

\usepackage{inconsolata}

\usepackage{graphicx}

\usepackage{amsmath}
\usepackage{makecell}
\usepackage{lipsum}
\usepackage{enumitem}
\usepackage{graphicx}
\usepackage{booktabs}
\usepackage{multirow}
\usepackage{CJK}
\usepackage{subcaption}
\usepackage{hyperref}
\usepackage{xparse}
\usepackage{pifont}
\usepackage[capitalize]{cleveref}
\usepackage{colortbl}
\usepackage{xcolor}

\usepackage{algorithm}
\usepackage{algpseudocode}
\usepackage{amsmath}

\title{Improving Cross-Format Robustness in Language Models with Multi-Format Training
}

\author{
  \textbf{June M. Liu}\textsuperscript{1}\thanks{Work done during an internship at Ant Group},
  \textbf{Shaomian Zheng}\textsuperscript{1},
  \textbf{He Cao}\textsuperscript{2},
  \textbf{Dingnan Jin}\textsuperscript{1},
  \textbf{Qing Cui}\textsuperscript{1},
  \textbf{Jun Zhou}\textsuperscript{1}\thanks{Corresponding author.\\Email: \href{mailto:jun.zhoujun@antgroup.com}{jun.zhoujun@antgroup.com}}
  \\
  \\
  \textsuperscript{1}Ant Group,
  \textsuperscript{2}International Digital Economy Academy (IDEA)
}

\begin{document}
\maketitle

\input{sec/0-abstract}
\input{sec/1-intro}
\input{sec/2-related}
\input{sec/3-data}
\input{sec/4-eval}
\input{sec/5-exp}
\input{sec/6-analysis}
\input{sec/7-conclusion}
\input{sec/8-Limitation}
\input{sec/9-Acknowledgement}



\bibliography{custom}

\appendix
\input{sec/appendix}

\end{document}

%% file: sec/0-abstract.tex
\begin{abstract}
Large language models often remain sensitive to answer format: a question solved correctly in one form may fail in another semantically equivalent form. 
To study this gap, we define cross-format robustness as the extent to which a model answers the same underlying question consistently across formats. 
We then compare full-format training with FormatMix, which expands only a subset of training items into multiple equivalent formats using either random or targeted selection. 
Across GLM4 and Llama-3.1, multi-format supervision consistently improves both task performance and cross-format robustness, whereas Multiple-choice question (MCQ)-only supervision alone brings little benefit and can even reduce robustness. 
We further find that expanding only about 30\% of the training set into multiple formats often recovers most of the gain from full-format training, and this effect appears across the model families and sizes we study. 
These results suggest that format diversity, rather than additional supervision alone, is the key driver of robustness. That lightweight multi-format augmentation is a practical way to make LLMs less sensitive to answer format without changing the base model.
\end{abstract}

%% file: sec/1-intro.tex
\section{Introduction}
Multiple-choice question (MCQ) answering is a standard benchmark for large language models (LLMs), but MCQ accuracy conflates two factors: whether a model has the relevant knowledge and whether it can use that knowledge under a particular question format. Recent work shows that LLMs are sensitive to answer options, choice presentation, and output format, often giving different answers to the same underlying question in multiple-choice and generative settings~\citep{li2024can,balepur2024artifacts,wang2025llms,balepur2025these,nguyen2025llms,takizawa2025mcqformatbench}. This matters beyond benchmark design. In real use, questions are rarely posed as MCQs; users ask open-ended questions, verify claims, or complete missing information. A model whose answers depend heavily on format is therefore not only hard to evaluate, but also hard to rely on.

\begin{figure}[!ht]
    \centering
    \small
    \vspace{-0.2cm}
    \includegraphics[width=\linewidth]{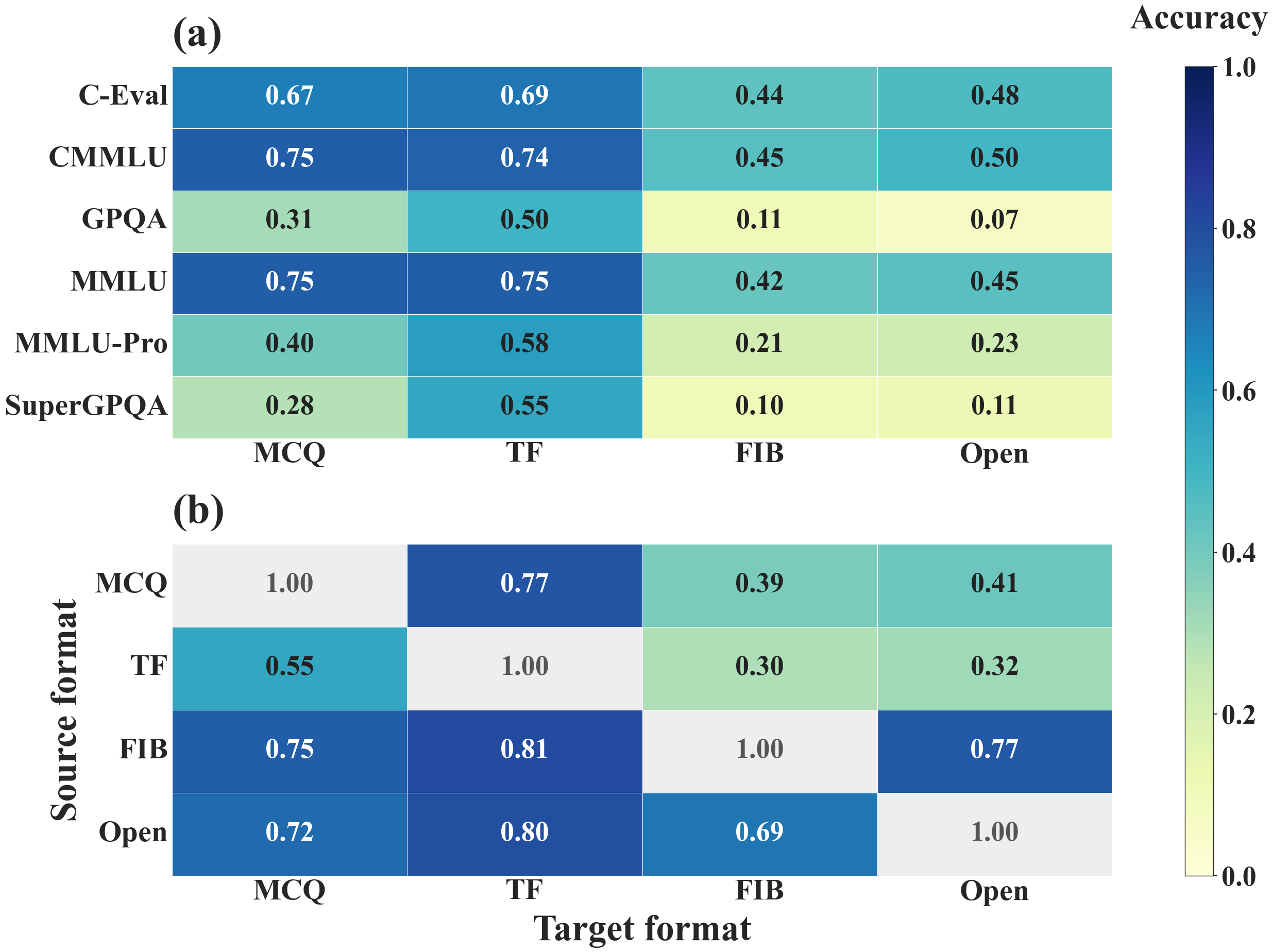}
    \vspace{-0.6cm}
    \caption{
    Cross-format inconsistency on semantically matched questions.
    (a) Accuracy of the same underlying items across four question formats on six benchmarks.
    (b) Conditional transfer accuracy: among items answered correctly in the source format, the fraction that remains correct in the target format.
    }
    \vspace{-0.4cm}
    \label{fig:format-inconsistency}
\end{figure}

In this paper, we study \emph{question-format robustness}: whether a model can answer semantically equivalent questions consistently across formats. To test this directly, we built a controlled multi-format setting. Starting from each original MCQ item, we construct semantically matched true/false (TF), fill-in-the-blank (FIB), and open-ended (OPEN) versions, with multiple paraphrases per format and validation through automatic filtering and human annotation. This design keeps the underlying content fixed while varying only the question format. It enables a simple diagnostic question: if a model answers an item correctly in one format, does that correctness transfer to the others? Figure~\ref{fig:format-inconsistency} shows that the answer is often no. Accuracy varies substantially across formats and benchmarks, and items solved in one format frequently fail in another.

Most prior work addresses this phenomenon from the evaluation side, for example, by adjusting MCQ scoring, perturbing answer choices, or improving answer matching~\citep{cavalin2025improving,chandak2025answer}. These methods are valuable for obtaining more reliable measurements, but they do not directly improve the model's behavior when the same content is presented outside MCQ form. We take a complementary perspective: question-format sensitivity is also a \emph{training} problem. The central question of this paper is whether multi-format supervision can make knowledge access more robust across semantically equivalent question formats.

To answer this question, we construct a multi-format QA corpus from several widely used MCQ benchmarks and introduce an item-level evaluation protocol that separates \emph{any-format solvability} from \emph{all-format consistency}. Specifically, $\mathrm{pass}@4$ measures whether an item is answered correctly in at least one of the four formats, whereas $\mathrm{pass}^{4}$ measures whether it is answered correctly in all four. Their ratio captures robustness conditioned on the item being solvable at all. We then compare standard MCQ-only fine-tuning with multi-format supervised fine-tuning. In addition to full expansion of every training item, we study budgeted settings that reformulate only a subset of items, either selected randomly or prioritized by observed cross-format inconsistency.

Our experiments show a clear pattern. Multi-format supervision substantially improves cross-format robustness, whereas matched-scale MCQ-only training mainly improves MCQ accuracy and leaves all-format consistency largely unchanged. On GLM4-9B, full multi-format SFT increases $\mathrm{pass}^{4}$ from 12.08\% to 19.61\% and improves robustness by 35\% relative to the baseline. Moreover, expanding only a subset of items recovers much of the gain, and inconsistency-based selection is more effective than random selection under the same expansion budget. These results suggest a practical lesson: format robustness is a trainable capability, and much of it can be gained without exhaustively rewriting the entire training set.
Our contributions are threefold:
\begin{itemize}[leftmargin=15pt, noitemsep]
    \item We reframe question-format sensitivity as a problem of \emph{robust knowledge access}, and construct a validated multi-format QA corpus with semantically matched MCQ, true/false, fill-in-the-blank, and open-ended versions.
    \item We introduce an item-level evaluation protocol for cross-format robustness based on $\mathrm{pass}@4$, $\mathrm{pass}^{4}$, and their ratio, which disentangles overall solvability from consistency across formats.
    \item We show that multi-format SFT improves cross-format robustness substantially beyond MCQ-only training, and that partial, inconsistency-targeted format expansion recovers most of the benefit of full multi-format supervision.
\end{itemize}

%% file: sec/2-related.tex
\section{Related Work}

Prior work most relevant to our study falls into three lines: evaluation of reliability across question-answering formats, model sensitivity to question format and prompt variation, and question reformulation as a semantic preservation problem.

\subsection{Evaluation Reliability Across Question-Answering Formats}
Multiple-choice evaluation is easy to score, but its results can reflect answer-set artifacts and scoring conventions rather than knowledge alone. Prior work shows that models can exploit the option set itself, that MCQ and long-form settings often probe different behaviors, and that success in MCQ may sometimes amount to selecting the least incorrect option rather than identifying a uniquely correct answer~\citep{balepur2024artifacts,li2024can,wang2025llms}. Accordingly, several studies propose consistency-aware scoring, altered answer choices, or improved extraction protocols to obtain more reliable estimates of model ability~\citep{cavalin2025improving,molfese2025right,balepur2025these,goral2025wait}.
Moving beyond MCQ can reduce option-induced shortcuts, and open-style questions or answer-matching protocols sometimes better align with human judgments~\citep{myrzakhan2024open,chandak2025answer,bernard2024equator}. However, this shifts the problem to answer scoring: LLM-based judges can be biased, adversarially vulnerable, and self-inconsistent~\citep{shi2025judging,raina2024llm,bavaresco2025llms,haldar2025rating}. Overall, these studies show that evaluation depends not only on question content, but also on the response format and scoring procedure.

\subsection{Format Sensitivity and Robustness}
Beyond answer-set artifacts, another line of work studies whether semantically equivalent prompts and output constraints yield stable model behavior. Nguyen et al. show that LLMs are biased toward particular output formats, and Takizawa et al. introduce MCQFormatBench to study robustness under variations within the MCQ family~\citep{nguyen2025llms,takizawa2025mcqformatbench}. More broadly, prompt sensitivity and stress-test studies show that small changes in phrasing can lead to substantial performance shifts~\citep{hua2025flaw,zhao2025stress}. Similar instability has also been observed in paraphrase consistency, value-laden questions, and knowledge representations that depend on superficial resemblance~\citep{moore2024large,haller2025llm}. Mechanistic analyses further suggest that answer-symbol prediction itself can be sensitive to the surface form used to elicit the response~\citep{wiegreffe2024answer}.
Our work builds on this literature but differs in two ways. First, we study format sensitivity in a controlled item-aligned setting by comparing semantically matched MCQ, true/false, fill-in-the-blank, and open-ended versions of the same question. Second, rather than treating inconsistency only as a diagnostic or evaluation concern, we ask whether it can be reduced through multi-format supervision.

\subsection{Question Reformulation}
Constructing aligned variants across formats is closely related to question rewriting. CANARD~\citep{elgohary2019can} established question rewriting as a semantic preservation problem, and subsequent works~\citep{fu2024qgeval,ye2022robustness} show that rewritten questions can differ in answerability, consistency, and hardness depending on the transformation used. 
These findings necessitate careful filtering and human validation to prevent meaning or difficulty shifts that could confound downstream comparisons.
Unlike prior work focused on rewrite quality, we treat reformulation as a controlled tool to study cross-format robustness and test whether multi-format supervision improves access to underlying knowledge.

\begin{figure*}[!ht]
    \centering
    \includegraphics[width=0.97\linewidth]{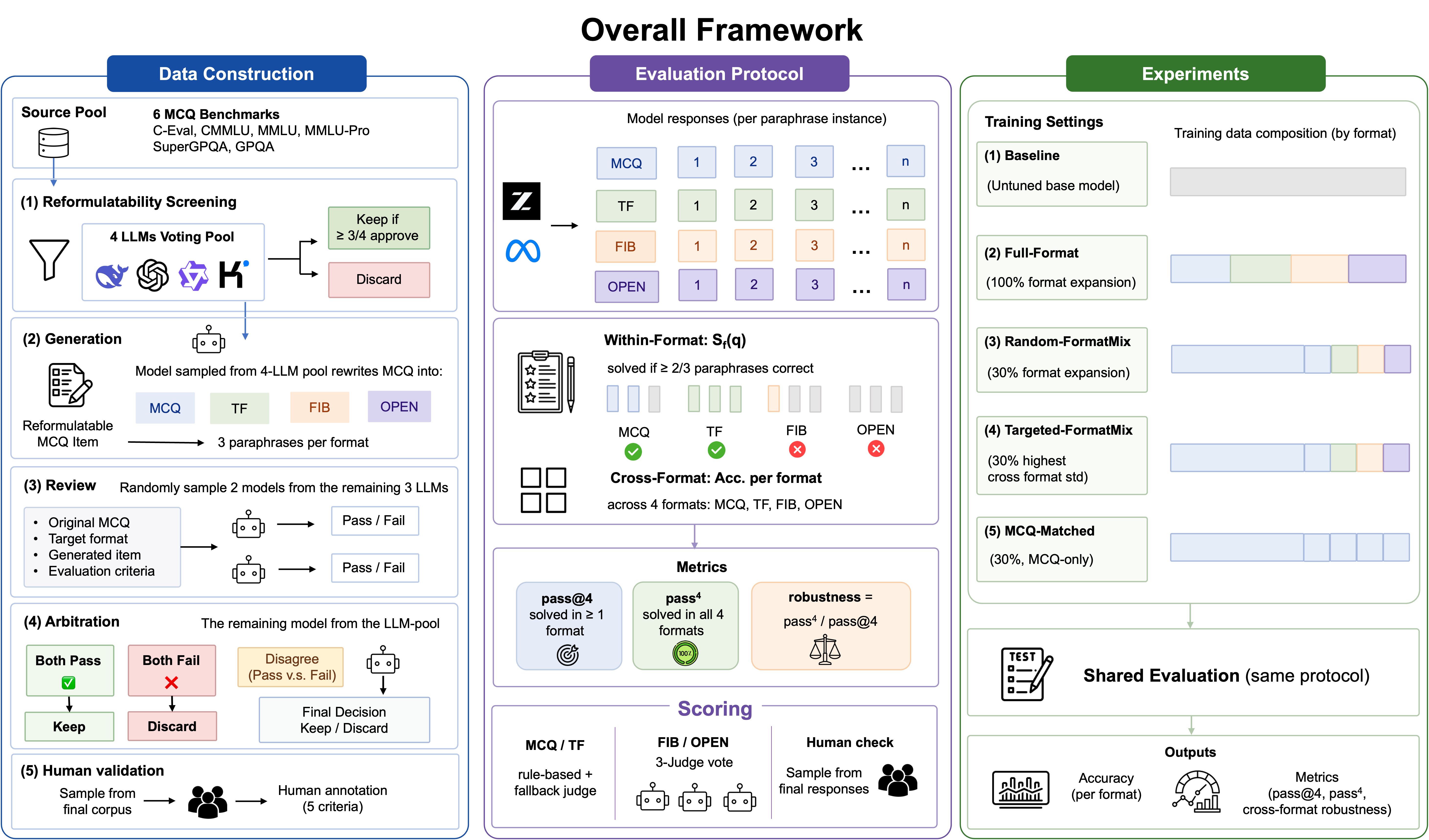}
    \vspace{-0.1cm}
    \caption{Overview of the proposed framework. Starting from source multiple-choice questions, we construct semantically aligned variants in four formats (MCQ, TF, FIB, and OPEN) through multi-stage rewriting and validation. We then evaluate models using an item-level cross-format protocol based on pass@4, pass$^{4}$, and robustness, and compare different multi-format supervision strategies.}
    \vspace{-0.3cm}
    \label{fig:framework}
\end{figure*}

%% file: sec/3-data.tex
\section{Data Construction}
\label{sec:data}

We construct an item-aligned multi-format QA corpus in four stages. We begin with 6 MCQ benchmarks, screen items for reformulatability, and retain only those that can be faithfully converted into non-MCQ formats and graded automatically. We then generate semantically aligned MCQ, true/false, fill-in-the-blank, and open-ended variants in the original language, with 3 paraphrases per format, followed by model-based review and arbitration. Human validation is conducted on a stratified subset. All train/test splits are defined at the source-item level before rewriting, so that no variants of the same item appear across splits.

\subsection{Source Benchmarks}

We collect source MCQ items from six widely used benchmarks: C-Eval~\citep{huang2023c}, CMMLU~\citep{li2024cmmlu}, MMLU~\citep{hendrycks2020measuring}, MMLU-Pro~\citep{wang2024mmlu}, GPQA~\citep{rein2023gpqa}, and SuperGPQA~\citep{du2026supergpqa}. These benchmarks cover a broad range of domains, languages, and difficulty levels, making them suitable source pools for constructing an item-aligned multi-format QA corpus. 
We split data at the source-item level before rewriting, so that all format variants and paraphrases of the same underlying item stay in the same split.

\subsection{Reformulatability Screening}

Since not all MCQ items can be faithfully converted into other question formats—some depend heavily on the presence of answer options, while others become ambiguous or difficult to score in other formats—we first perform a reformulatability screening step. For each source item, four LLMs independently assess whether it is suitable for multi-format reformulation: DeepSeek-V3.1~\citep{deepseekai2024deepseekv3technicalreport}, Qwen3-235B-A22B-Instruct-2507~\citep{qwen3technicalreport}, Kimi-K2.5~\citep{kimiteam2026kimik25visualagentic}, and GPT-OSS-120B~\citep{openai2025gptoss120bgptoss20bmodel}. We retain an item only if at least three of the four models approve it under a strict rubric focused on answer uniqueness, option independence, and automatic gradability. The full screening prompt is provided in Appendix~\ref{fig:appendix-filtering-prompt}.

\subsection{Multi-Format Rewriting}
\label{sec:multi-format-rewriting}

For each retained source item, we construct semantically aligned variants in four formats: MCQ, true/false (TF), fill-in-the-blank (FIB), and open-ended (OPEN). We preserve the source language of the original benchmark and generate three paraphrases per format with temperature $=1.0$ and top-$p=0.95$. To reduce dependence on any single model family, we adopt a multi-model generation, review, and arbitration pipeline inspired by \citet{gao2025strategic}. Specifically, we utilize the same four-model pool used in the screening stage. For each generation instance, we randomly select one model from this pool to act as the rewriter. The rewrites are designed to preserve the underlying factual content and gold decision of the source item while changing only the interaction format and response space. For MCQ, we paraphrase the original question while preserving the answer set and correct answer, allowing the option order to vary. For TF, we convert the item into a single proposition derived from one source option: propositions based on the original correct option are labeled \texttt{True}, while propositions based on incorrect options are labeled \texttt{False}. For FIB, we rewrite the item into a cloze-style question, and for OPEN, we rewrite the same content into an open-ended question that must be answered without access to options. Because downstream scoring for non-discrete formats relies on LLM-based judgment rather than exact string matching, we do not impose additional canonicalization constraints on FIB or OPEN answers at construction time.

We then apply format-specific quality control to all generated candidates. For each candidate, we randomly sample two models from the remaining three LLMs in the pool to serve as independent reviewers. Each rewrite is reviewed for meaning preservation, answer preservation, self-containedness, format fidelity, and scorability. If both reviewers approve the rewrite, it is retained; if both fail it, it is discarded. In cases of reviewer disagreement (one pass and one fail), the candidate enters an arbitration step. The fourth and final remaining model in the pool acts as the arbitrator, evaluating the candidate alongside the scores and reasoning provided by the two reviewers to make a final keep or discard decision. To reduce answer-label prior artifacts, we encourage balanced label distributions during generation and enforce them at the corpus level after generation. Specifically, the prompts instruct the model to randomize MCQ correct-option positions and TF truth values whenever possible, and we further monitor the aggregate distribution of generated labels; if particular MCQ option positions or TF labels are overrepresented, affected items are rewritten until the final corpus is approximately balanced. Detailed generation, review, and arbitration prompts are provided in Appendix~\ref{app:prompt-templates}, and the review rubric is shown in Table~\ref{tab:rewrite-rubric}. After screening and rewriting, we retain 11{,}842 source items. Since each item is associated with four formats and three paraphrase variants per format, the final corpus contains 142{,}104 item-format-paraphrase instances. See Table~\ref{tab:source-screening} for detailed statistics and Algorithm~\ref{alg:multi_format} for overall data construction pipeline.

\subsection{Human Validation of Rewrites}

To assess rewrite quality, we conduct human validation on a subset of the generated data. Three independent annotators judge each rewrite on five binary criteria: meaning preservation, answer preservation, self-containedness, clarity and naturalness, and scorability. The full rubric is provided in Appendix~\ref{app:rewrite-rubric}.
Table~\ref{tab:rewrite-human-results} shows that the rewrites are high quality overall. Average scores exceed 90\% on all five criteria, including 97.83\% for meaning preservation and 91.67\% for answer preservation, indicating that most reformulations remain semantically aligned with the source item and preserve the intended answer. Clarity/naturalness (99.67\%) and scorability (98.83\%) are near ceiling, while self-containedness is somewhat lower at 90.67\%, reflecting the greater difficulty of preserving full context across formats. Full agreement is also high, especially for clarity/naturalness (98.98\%) and scorability (96.43\%). Overall, these results support the reliability of the rewritten corpus for controlled cross-format evaluation.

%% file: sec/4-eval.tex
\section{Evaluation Protocol}
\label{sec:evaluation-protocol}

Our goal is to measure not only whether a model can answer a question, but whether that answer remains accessible when the same underlying item is presented in different formats. Marginal accuracies alone are insufficient for this purpose: two models can achieve similar per-format accuracy while succeeding on entirely different subsets of items. To isolate cross-format robustness, we evaluate at the source-item level and aggregate in two stages: first across paraphrases within a format, and then across formats for the same item.

\subsection{Item-Level Aggregation and Cross-Format Metrics}
Each source item $q$ is instantiated in four formats,
$f \in \{\mathrm{MCQ}, \mathrm{TF}, \mathrm{FIB}, \mathrm{OPEN}\}$,
with three paraphrases per format.
For a given item and format, we say the model \emph{solves} that format if it answers at least two of the three paraphrases correctly:
{
\setlength{\abovedisplayskip}{1pt}
\setlength{\belowdisplayskip}{1pt}
\setlength{\abovedisplayshortskip}{1pt}
\setlength{\belowdisplayshortskip}{1pt}
\[
s_f(q)=\mathbb{I}\left[\sum_{v=1}^{3}\mathrm{correct}(q,f,v)\ge 2\right].
\]
}
This majority-over-paraphrases rule reduces sensitivity to spurious wording while still requiring stable success within a format. Per-format accuracies reported in our results are the averages of $s_f(q)$ over items.
We then report three item-level metrics.
\textbf{pass@4} measures \emph{any-format solvability}: the fraction of items solved in at least one of the four formats,
{
\setlength{\abovedisplayskip}{1pt}
\setlength{\belowdisplayskip}{1pt}
\setlength{\abovedisplayshortskip}{1pt}
\setlength{\belowdisplayshortskip}{1pt}
\begin{equation}
\mathrm{pass@4}=
\frac{1}{|\mathcal{Q}|}
\sum_{q\in\mathcal{Q}}
\mathbb{I}\!\left[\sum_{f\in\mathcal{F}} s_f(q)\ge 1\right].
\end{equation}
\textbf{pass$^{4}$} measures \emph{all-format consistency}: the fraction of items solved in all four formats,
\begin{equation}
\mathrm{pass}^{4}=
\frac{1}{|\mathcal{Q}|}
\sum_{q\in\mathcal{Q}}
\mathbb{I}\!\left[\sum_{f\in\mathcal{F}} s_f(q)=4\right].
\end{equation}
}
Finally, we define \textbf{cross-format robustness} as
{
\setlength{\abovedisplayskip}{1pt}
\setlength{\belowdisplayskip}{1pt}
\setlength{\abovedisplayshortskip}{1pt}
\setlength{\belowdisplayshortskip}{1pt}
\[
\begin{aligned}
\frac{\mathrm{pass}^{4}}{\mathrm{pass@4}}
={}& P\bigl(\text{solved in all four formats} \mid {} \\
&\qquad \text{solved in at least one format}\bigr).
\end{aligned}
\]
}
This conditioning is important because it filters out unsolvable items, allowing us to measure robustness solely among questions the model is capable of answering.
In other words, $\mathrm{pass@4}$ asks whether the model can access the answer under \emph{some} interface, whereas $\mathrm{pass}^{4}$ asks whether that access is preserved across \emph{all} interfaces.

\subsection{Automatic Scoring}
We use a deterministic scoring rule available for each format.
For MCQ and TF, we first apply rule-based extraction to recover the model's final option letter or true/false judgment.
If no unambiguous final answer can be extracted, or if the output contains multiple conflicting answers, we fall back to LLM-as-a-Judge scoring.
For FIB and OPEN, surface-form matching is often too brittle: a response may be semantically correct even when it differs from the reference answer lexically.
We therefore score these formats directly with LLM judges.
Our judge panel consists of GPT-5, Gemini-2.5-Pro, and Claude-Sonnet-4.6.
Each judge receives the question, the reference answer, and the model response, and outputs a binary correctness label.
We use the majority vote across the three judges as the final automatic score.

\subsection{Human Validation of Automatic Scoring}

To assess the reliability of the scoring pipeline, we manually annotate a sample of 200 model responses and compare the automatic labels against the majority human label.
Agreement between human annotation and automatic scoring is 95.33\%.
Human annotators also show high internal consistency, with 87.76\% three-way exact agreement and 91.90\% pairwise agreement.
These results suggest that the remaining scoring ambiguity is limited and that the automatic pipeline is reliable enough for large-scale evaluation.

%% file: sec/5-exp.tex
\section{Experiments}
\label{sec:experiments}

Our item-aligned corpus allows controlled training: for each question, models see either one canonical MCQ or multiple matched formats. This tests whether multi-format supervision improves both per-format accuracy and cross-format consistency.

\subsection{Training Regimes}
\label{sec:training-conditions}

We structure the main experiments around three questions:
(Q1) Does multi-format supervision improve cross-format robustness?
(Q2) Are any gains due to format diversity rather than simply more supervised training?
(Q3) Under a fixed expansion budget, is it better to expand items at random or to prioritize those that are most format-sensitive under the base model?

All conditions are built from the same set of source training items and differ only in how those items are instantiated during training. A source item kept in MCQ-only form contributes its three MCQ paraphrases, whereas a source item expanded into all four formats contributes 12 instances (4 formats $\times$ 3 paraphrases). If a fraction $\rho$ of source items is expanded, the resulting training set contains $(1+3\rho)$ times as many instances as MCQ-only training. In the main table, we use $\rho=0.3$ as a representative mid-budget setting, yielding a 1.9$\times$ training set relative to MCQ-only supervision, compared with 4.0$\times$ for expanding every item. We do not treat 30\% as an optimized choice; Section~\ref{sec:analysis} later reports a 10/20/30\% sweep.

\noindent
(1) \textbf{Untuned base model (Baseline).}
The base model is evaluated without fine-tuning. This establishes the model's native level of cross-format robustness.

\noindent
(2) \textbf{Full-Format SFT.}
Every training item is expanded into all four formats. This condition represents exhaustive multi-format supervision and serves as an upper-bound condition rather than a budget-matched comparison.

\noindent
(3) \textbf{Random-FormatMix SFT (30\%).}
A fixed 30\% of source training items are randomly selected and expanded into all four formats, while the remaining 70\% remain MCQ-only. This condition tests whether partial multi-format expansion already captures a substantial fraction of the benefit of full expansion.

\noindent
(4) \textbf{Targeted-FormatMix SFT (30\%).}
The same 30\% expansion budget is used, but items are selected according to their format sensitivity under the untuned base model on the training split only. For each source item $q$, we compute
\[
\mathrm{Cross\text{-}format\ Std}(q)=
\mathrm{Std}_{f\in\mathcal{F}}
\bigl(\mathrm{acc}(q,f)\bigr),
\]
where $\mathcal{F}=\{\mathrm{MCQ}, \mathrm{TF}, \mathrm{FIB}, \mathrm{OPEN}\}$ and $\mathrm{acc}(q,f)$ is the mean correctness over the three paraphrases in format $f$. A higher level indicates that the item is accessible under some interfaces but not others, making it a natural target for robustness-oriented supervision. Compared with binary item-level indicators such as pass@4 or pass$^{4}$, this continuous score provides a finer-grained ranking signal for data selection.

\noindent
(5) \textbf{Size-Matched MCQ-only SFT (MCQ-Matched).}
This control matches the total number of training instances of the 30\% FormatMix conditions, but keeps all supervision in MCQ form through resampling. It isolates the effect of format diversity from the effects of additional supervision and longer training.

\subsection{Training Details}
\label{sec:training-details}

Our main experiments use GLM4-9B as the base model. We perform full-parameter supervised fine-tuning for one epoch with learning rate $1\times10^{-6}$, cosine learning rate decay, and warmup ratio 0.1. Training is conducted on 8 H20 GPUs. We use batch size 2 per device and gradient accumulation 4, resulting in an effective batch size of 64. 
All SFT runs use the same optimizer and data split. Random-FormatMix, Targeted-FormatMix, and MCQ-Matched are matched in total training instances (30\% budget), forming our main controlled comparison. Full-Format SFT uses more data and serves as an upper bound.

\begin{table*}[!ht]
\centering
\small
\setlength{\tabcolsep}{5pt}
\renewcommand{\arraystretch}{1.1}
\begin{tabular}{lccccccc}
\toprule
\textbf{Condition} 
& \textbf{pass@4} 
& \textbf{$\mathrm{pass}^4$} 
& \textbf{\shortstack{Cross-format\\robustness}} 
& \textbf{\shortstack{MCQ\\Acc.}} 
& \textbf{\shortstack{TF\\Acc.}} 
& \textbf{\shortstack{FIB\\Acc.}} 
& \textbf{\shortstack{OPEN\\Acc.}} \\
\midrule
Baseline 
& 0.7669 
& 0.1208 
& 0.1575 
& 0.4769 
& 0.6255 
& 0.2478 
& 0.2660 \\

Full-Format 
& \textbf{0.9222} 
& \textbf{0.1961} 
& \textbf{0.2126} 
& \underline{0.5265} 
& \textbf{0.7747} 
& \textbf{0.3253} 
& \textbf{0.3421} \\

Random-FormatMix
& \underline{0.9041} 
& 0.1695 
& 0.1875 
& 0.5156 
& \underline{0.7392} 
& 0.2798 
& 0.3066 \\

Targeted-FormatMix
& 0.9015 
& \underline{0.1775} 
& \underline{0.1969} 
& 0.5111 
& \underline{0.7392} 
& \underline{0.2842} 
& \underline{0.3071} \\

MCQ-Matched
& 0.8157 
& 0.1209 
& 0.1482 
& \textbf{0.5320} 
& 0.6354 
& 0.2261 
& 0.2657 \\
\bottomrule
\end{tabular}
\vspace{-0.3cm}
\caption{Main results on GLM4-9B. Under the matched 30\% budget, both FormatMix variants outperform the size-matched MCQ-only control, showing that gains come from format diversity rather than training volume alone. Targeted-FormatMix SFT is better on pass$^{4}$ and robustness, while Random-FormatMix SFT is slightly better on pass@4. Bold and underlined values denote the best and second-best results.}
\vspace{-0.4cm}
\label{tab:main-results}
\end{table*}

\subsection{Main Results}
\label{sec:main-results}

\paragraph{Full multi-format supervision substantially improves cross-format robustness.} As shown in Table~\ref{tab:main-results}, compared with the baseline, Full-Format SFT improves accuracy in every format and yields substantial gains on the three metrics: pass@4 increases by about 20\%, pass$^{4}$ by about 62\%, and cross-format robustness by 35\%. In addition, there is a larger percentage of gains on TF, FIB, and OPEN. 
This pattern suggests that multi-format supervision does more than reinforce benchmark-specific MCQ behavior: it makes the underlying answer more consistently accessible across interfaces.

\paragraph{The gain is not explained by training volume alone.} The Size-Matched MCQ-only SFT control improves MCQ performance (0.4769 to 0.5320) but leaves pass$^{4}$ essentially unchanged (0.1208 to 0.1209) and slightly reduces cross-format robustness (0.1575 to 0.1482). In other words, additional single-format supervision sharpens performance in the seen interface without improving interface-invariant access to the answer. The contrast with the budget-matched FormatMix SFT conditions shows that the key ingredient is format diversity rather than simply more supervised updates.

\paragraph{Partial expansion already captures most of the benefit, and targeted expansion favors consistency over breadth.} 
Under the fixed 30\% expansion budget, both Random-FormatMix and Targeted-FormatMix SFT substantially outperform the size-matched MCQ-only control on all cross-format metrics. Relative to the gains of Full-Format SFT over the base model, Random-FormatMix SFT recovers 88\% of the pass@4 improvement, while Targeted-FormatMix SFT recovers 75\% of the pass$^{4}$ improvement and 72\% of the robustness improvement, despite using less than half the total training volume of Full-Format SFT (1.9$\times$ vs.\ 4.0$\times$ MCQ-only training size). The two partial-expansion strategies also differ in a meaningful way: Random-FormatMix SFT is slightly stronger on pass@4, whereas Targeted-FormatMix SFT is numerically stronger on pass$^{4}$ and cross-format robustness. This is consistent with our selection criterion: random expansion broadens coverage, while targeted expansion preferentially repairs items whose answer accessibility is most format-sensitive.

%% file: sec/6-analysis.tex
\section{Analysis}
\label{sec:analysis}

We now address three questions: the required scale of multi-format expansion, the behavioral changes induced by supervision, and whether these patterns generalize across model scales and families.

\subsection{Multi-Format Expansion: Budget Efficiency \& Cross-Family Transfer}

\begin{table}[!ht]
\centering
\small
\vspace{-0.2cm}
\setlength{\tabcolsep}{5pt}
\renewcommand{\arraystretch}{1.1}
\resizebox{0.95\columnwidth}{!}{
\begin{tabular}{lccc}
\toprule
\textbf{Condition} 
& \textbf{pass@4} 
& \textbf{$\mathrm{pass}^4$} 
& \textbf{\shortstack{Cross-format \\Robustness}} \\
\midrule
Baseline                 & 0.7669 & 0.1208 & 0.1575 \\
Random 10\%    & \textbf{0.9074} & 0.1615 & 0.1779 \\
Random 20\%    & 0.9053 & 0.1648 & 0.1821 \\
Random 30\%    & 0.9041 & 0.1695 & 0.1875 \\
Targeted 10\%  & 0.8800 & 0.1682 & 0.1912 \\
Targeted 20\%  & 0.8779 & \textbf{0.1775} & \textbf{0.2022} \\
Targeted 30\%  & 0.9015 & \textbf{0.1775} & 0.1969 \\
\bottomrule
\end{tabular}
}
\vspace{-0.2cm}
\caption{Budget efficiency of partial multi-format expansion on GLM4-9B. 
Targeted selection consistently trades some breadth for better cross-format consistency.}
\label{tab:formatmix}
\end{table}

\begin{table}[!ht]
\centering
\small
\setlength{\tabcolsep}{5pt}
\renewcommand{\arraystretch}{1.1}
\vspace{-0.5cm}
\resizebox{\linewidth}{!}{
\begin{tabular}{lccc}
\toprule
\textbf{Model} 
& \textbf{pass@4} 
& \textbf{$\mathrm{pass}^4$} 
& \textbf{\shortstack{Cross-format\\robustness}} \\
\midrule
Llama-3.1-70B (ref.) 
& 0.7401 & 0.1534 & 0.2073 \\
\midrule
Baseline (Llama-3.1-8B) 
& 0.6713 & 0.0669 & 0.0996 \\

Full-Format
& \textbf{0.9011} & \textbf{0.1344} & \textbf{0.1492} \\

MCQ-Matched 
& 0.7502 & 0.0680 & 0.0907 \\

Random 10\% 
& 0.8588 & 0.1162 & 0.1353 \\

Random 20\% 
& 0.8783 & 0.1221 & 0.1391 \\

Random 30\% 
& 0.8897 & 0.1302 & 0.1463 \\

Targeted 10\% 
& 0.8724 & 0.1200 & 0.1376 \\

Targeted 20\% 
& 0.8990 & 0.1268 & 0.1410 \\

Targeted 30\% 
& 0.8500 & 0.1192 & 0.1402 \\
\bottomrule
\end{tabular}
}
\vspace{-0.2cm}
\caption{Results on the Llama family. Llama-3.1-70B is included as a larger-scale reference model, while all SFT variants are based on Llama-3.1-8B. 
}
\label{tab:llama}
\vspace{-0.3cm}
\end{table}

Table~\ref{tab:formatmix} shows that even limited multi-format expansion is highly effective. On GLM4-9B, expanding only 10\% of source items already raises pass@4 from 0.7669 to 0.9074 under random selection. However, pass$^{4}$ and cross-format robustness continue to improve as the expansion ratio increases from 10\% to 30\%, suggesting that a small expansion budget is sufficient to make many items solvable in at least one format, while stronger cross-format consistency requires additional multi-format supervision.
Under the same budget, Random-FormatMix SFT is consistently stronger on pass@4, whereas Targeted-FormatMix SFT is stronger on pass$^{4}$ and cross-format robustness. This suggests a trade-off between broader coverage and greater within-item stability: random expansion exposes the model to more items, while targeted expansion focuses supervision on questions that are most format-sensitive.

The same high-level pattern largely holds for Llama-3.1-8B (Table~\ref{tab:llama}). Partial multi-format expansion again recovers most of the benefit of Full-Format SFT, indicating that exhaustive expansion is unnecessary. However, the advantage of targeted selection is less stable than on GLM4-9B, since random and targeted expansion perform similarly overall, and the best random setting is slightly stronger on the consistency metrics. We therefore view targeted selection as a useful budget-allocation heuristic rather than a universally superior strategy.

\subsection{Does Model Scale Affect the Results?}

Scaling already brings a substantial amount of native format robustness. As shown in Table~\ref{tab:scale}, the GLM4-32B baseline exceeds the GLM4-9B baseline on all three metrics, with especially large gains on pass$^{4}$ and cross-format robustness. Indeed, the 32B baseline robustness (0.2890) is already higher than the 9B model after Full-Format SFT (0.2126), indicating that larger models internalize more interface invariance even before task-specific adaptation.
Multi-format SFT nevertheless remains useful at 32B. Full-Format SFT further improves the 32B model from 0.8994 to 0.9332 on pass@4, from 0.2599 to 0.3111 on pass$^{4}$, and from 0.2890 to 0.3333 on cross-format robustness. Targeted-FormatMix also yields a smaller but still positive shift. The key pattern is therefore one of diminishing headroom, not redundancy: model scale and multi-format SFT are complementary, but the marginal gain of supervision becomes smaller as the base model becomes more inherently robust.

\begin{table}[!ht]
\centering
\small
\setlength{\tabcolsep}{5pt}
\renewcommand{\arraystretch}{1.1}
\begin{tabular}{lccc}
\toprule
\textbf{Condition} 
& \textbf{pass@4} 
& \textbf{$\mathrm{pass}^4$} 
& \textbf{\shortstack{Cross-format\\robustness}} \\
\midrule
GLM4-9B Baseline   & 0.7669 & 0.1208 & 0.1575 \\
GLM4-9B Full       & 0.9222 & 0.1961 & 0.2126 \\
GLM4-9B Targeted   & 0.9015 & 0.1775 & 0.1969 \\
\midrule
GLM4-32B Baseline  & 0.8994 & 0.2599 & 0.2890 \\
GLM4-32B Full      & 0.9332 & 0.3111 & 0.3333 \\
GLM4-32B Targeted  & 0.9091 & 0.2798 & 0.3078 \\
\bottomrule
\end{tabular}
\vspace{-0.2cm}
\caption{Effect of model scale.}
\vspace{-0.3cm}
\label{tab:scale}
\end{table}

\subsection{Validation on External Dataset}

To examine whether our main conclusions also transfer beyond the constructed benchmark, we evaluate on the BIG-Bench Hard (BBH)~\citep{suzgun2023challenging} covering MCQ, OPEN, and TF tasks. Since BBH does not provide semantically aligned reformulations of the same underlying item across formats, item-level metrics such as pass@4 and pass$^4$ are not applicable here, so we report per-format accuracy only. As shown in Table~\ref{tab:bbh_external}, the overall pattern is broadly consistent with our in-domain results: Full-Format SFT improves over the baseline on all three formats, with gains on MCQ (0.5430$\rightarrow$0.5523), OPEN (0.2340$\rightarrow$0.2460), and TF (0.6214$\rightarrow$0.6228). Targeted-FormatMix remains the strongest 30\%-budget variant and is especially competitive on OPEN, where it nearly matches Full-Format SFT (0.2450 vs.\ 0.2460). By contrast, MCQ-Matched achieves the best results on the two closed-form formats (0.5596 on MCQ and 0.6263 on TF) but lags behind on OPEN (0.2230), suggesting that additional single-format supervision primarily benefits constrained answer spaces, whereas format-diverse supervision transfers more clearly to open-ended settings.

\begin{table}[!ht]
\centering
\small
\vspace{-0.3cm}
\resizebox{\columnwidth}{!}{
\begin{tabular}{lccc}
\toprule
Condition & MCQ Acc. & TF Acc. & OPEN Acc. \\
\midrule
Baseline (GLM4-9B) & 0.5430 & 0.6214 & 0.2340 \\
Full-Format & \underline{0.5523} & \underline{0.6228} & \textbf{0.2460} \\
Random-FormatMix & 0.5506 & 0.6068 & 0.2350 \\
Targeted-FormatMix & 0.5493 & 0.6054 & \underline{0.2450} \\
MCQ-Matched & \textbf{0.5596} & \textbf{0.6263} & 0.2230 \\
\bottomrule
\end{tabular}
}
\vspace{-0.3cm}
\caption{External validation on BIG-Bench Hard.}
\vspace{-0.6cm}
\label{tab:bbh_external}
\end{table}

%% file: sec/7-conclusion.tex

\section{Conclusion}
We study a simple but important robustness question: whether the same knowledge remains accessible when a question is expressed in different answer formats. Using an item-aligned multi-format benchmark and evaluation protocol, strong LLMs remain sensitive to format changes. Across GLM4 and Llama-3.1, multi-format supervision improves solvability and consistency, and these gains cannot be explained by additional MCQ-only supervision alone. We further find that much of the benefit can be obtained with partial expansion, while larger models, although more robust, still exhibit a clear gap between one-format success and consistent success across formats. 

%% file: sec/8-Limitation.tex
\clearpage
\section*{Limitations}
This study has several limitations. First, we evaluate FormatMix on a fixed set of four text-based answer formats and on two model families, so the findings may not directly generalize to multimodal settings, freer-form interactive generation, or other architectures.
Moreover, our corpus construction pipeline applies strict reformulatability filtering to ensure semantic alignment and reliable automatic evaluation across formats. While this improves experimental control, extending the framework to a broader range of naturally occurring questions remains an important direction for future work.
Finally, evaluation for FIB and OPEN formats relies primarily on LLM-as-a-judge scoring. Although human validation shows high agreement with automatic evaluation, incorporating larger-scale human assessment or open evaluation protocols would further strengthen reliability and reproducibility.


%% file: sec/9-Acknowledgement.tex
\section*{Acknowledgments}
This work was supported by Ant Group Research Intern Program.

%% file: sec/appendix.tex
\appendix
\raggedbottom
\section{Appendix}
\label{sec:appendix}

\subsection{Human Validation Rubric and Results of Rewrites}
\label{app:rewrite-rubric}
We conduct human validation to assess the quality of the rewritten multi-format questions. Each rewritten question is evaluated along five binary dimensions: meaning preservation, answer preservation, self-containedness, clarity and naturalness, and scorability. Table~\ref{tab:rewrite-rubric} presents the annotation rubric. Table~\ref{tab:rewrite-human-results} summarizes the results. Avg. Score is the mean binary label across annotations, and All-Positive Agreement measures the fraction of items for which all three annotators assign a positive label. These manual checks provide an independent lower-bound estimate of rewrite quality and confirm that the automatic filtering pipeline removes most problematic cases before data release.

\begin{table*}[!ht]
\centering
\small
\renewcommand{\arraystretch}{1.15}
\begin{tabular}{@{}p{0.17\textwidth}p{0.38\textwidth}p{0.38\textwidth}@{}}
\toprule
\textbf{Criterion} & \textbf{Score 1} & \textbf{Score 0} \\
\midrule
Meaning preservation
& The rewritten question asks the same underlying question as the original item. Key entities, conditions, temporal scope, comparison relations, and negation are preserved.
& The rewrite changes the core meaning, such as replacing the target entity, adding or removing key conditions, changing the comparison relation, narrowing or broadening the scope, or flipping the polarity. \\

Answer preservation
& The rewritten reference answer is compatible with the rewritten question and does not obviously change the intended answer.
& The rewritten question and reference answer are clearly inconsistent, or the rewrite makes the original answer no longer valid. \\

Self-containedness
& The rewritten question can be understood and answered on its own, without relying on the original options, previous context, images, tables, or unresolved references.
& The question depends on missing context, such as ``the above options'', ``the figure'', ``this statement'', or other unresolved references. \\

Clarity and naturalness
& The expression is fluent and clear, with no obvious grammatical errors, broken sentences, corrupted symbols, or confusing references.
& The rewrite is unclear or unnatural, contains serious grammatical issues, incomplete sentences, corrupted text, or ambiguous references. \\

Scorability
& The question has a stable and explicit correctness criterion, and the reference answer is sufficient for grading.
& The question is too open-ended or ambiguous, admits many substantially different valid answers, or has an insufficiently specific reference answer. \\
\bottomrule
\end{tabular}
\caption{Human validation rubric for rewritten multi-format questions. Each dimension is annotated as a binary score.}
\label{tab:rewrite-rubric}
\end{table*}

\begin{table}[!ht]
\centering
\small
\renewcommand{\arraystretch}{1.12}
\resizebox{\linewidth}{!}{
\begin{tabular}{@{}lcc@{}}
\toprule
\textbf{Criterion} & \textbf{Avg. Score} & \textbf{All-Positive Agreement} \\
\midrule
Meaning preservation      & 97.83\% & 93.88\% \\
Answer preservation       & 91.67\% & 81.12\% \\
Self-containedness        & 90.67\% & 73.47\% \\
Clarity and naturalness   & 99.67\% & 98.98\% \\
Scorability               & 98.83\% & 96.43\% \\
\bottomrule
\end{tabular}
}
\caption{
Human validation results for rewritten questions. Metrics: Avg. Score (mean binary rating) and All‑Positive Agreement (full consensus). Rewrites perform strongly across all dimensions, especially clarity/naturalness and scorability.
}
\label{tab:rewrite-human-results}
\end{table}

\subsection{Source Benchmark and Screening Statistics}

\paragraph{C-Eval}
C-Eval~\cite{huang2023c} is a comprehensive Chinese evaluation suite designed to assess the advanced knowledge and reasoning abilities of foundation models in a Chinese linguistic context. The dataset contains 13,948 multiple-choice questions meticulously collected from real-world human examinations. It covers a broad scope of 52 diverse disciplines, ranging from humanities to science and engineering. The benchmark evaluates models across four distinct difficulty levels: middle school, high school, college, and professional, establishing it as a standard and highly challenging metric for assessing Chinese language proficiency and domain-specific knowledge.

\paragraph{CMMLU}
CMMLU~\cite{li2024cmmlu} (Chinese Massive Multitask Language Understanding) is an extensive benchmark designed to measure the knowledge and reasoning capabilities of large language models within the Chinese cultural and linguistic context. Sourced from various professional assessments and academic exams, it consists of 11,528 multiple-choice questions. The benchmark covers 67 disciplines spanning natural sciences, social sciences, engineering, and humanities. Notably, it includes numerous China-specific subjects that require deep regional understanding. The difficulty level comprehensively ranges from primary school all the way to advanced professional tiers.

\paragraph{MMLU}
MMLU~\cite{hendrycks2020measuring} (Massive Multitask Language Understanding) is a widely adopted English benchmark created to evaluate the encyclopedic knowledge and zero-shot/few-shot capabilities of language models. The dataset consists of 15,908 multiple-choice questions sourced from human examinations such as the GRE, USMLE, and various Advanced Placement (AP) tests. It spans 57 subjects across STEM, humanities, social sciences, and applied professional fields. The difficulty of the questions covers a wide spectrum from elementary school to expert professional levels, making it a foundational standard for general knowledge evaluation.

\paragraph{MMLU-Pro}
MMLU-Pro~\cite{wang2024mmlu} is a more robust and challenging enhancement of the original MMLU benchmark, specifically tailored to evaluate deep reasoning capabilities rather than mere memorization. It comprises over 12,000 rigorously curated questions sourced from a combination of the original MMLU dataset, college-level science exams (e.g., SciBench), and advanced STEM repositories (e.g., TheoremQA). The dataset condenses its focus into 14 distinct domains (including Biology, Chemistry, Physics, Law, and Math). Unlike the standard four-option format, MMLU-Pro increases the number of distractors to provide ten options per multiple-choice question, dramatically reducing the probability of random guessing. Targeted at the graduate and professional levels, it is significantly more difficult than MMLU, typically causing a 16\% to 33\% accuracy drop in state-of-the-art models.

\paragraph{GPQA}
GPQA~\cite{rein2023gpqa} (Graduate-Level Google-Proof Q\&A) is an exceptionally challenging benchmark designed to test expert-level scientific reasoning. It consists of 448 high-quality multiple-choice questions exclusively written and verified by domain experts (PhDs and PhD candidates). The dataset is strictly focused on three core domains: Biology, Chemistry, and Physics. The questions are intentionally designed to be ``Google-proof,'' meaning that highly skilled non-expert human validators equipped with unrestricted internet access achieve only around 34\% accuracy. The benchmark represents the extreme upper bound of graduate-level difficulty, where even domain experts with PhDs achieve approximately 65\% accuracy without external aids.

\paragraph{SuperGPQA}
SuperGPQA~\cite{du2026supergpqa} is a massive expansion of the graduate-level evaluation paradigm, designed to measure model capabilities across a vastly broader spectrum of human knowledge. Constructed through a novel Human-LLM collaborative filtering mechanism involving over 80 expert annotators, it contains 26,529 multiple-choice questions. The dataset scales the evaluation to 285 graduate-level disciplines, encompassing not only mainstream STEM fields but also heavily overlooked long-tail domains such as light industry, agriculture, and specialized services. With a heavy emphasis on mathematical calculation and formal reasoning, the difficulty remains strictly at the graduate level, serving as a challenging frontier benchmark where even the most advanced contemporary reasoning models struggle to significantly surpass a 60\% to 65\% accuracy threshold.

The aforementioned six benchmarks collectively serve as the foundational data pool for constructing our dataset. We provide the details of aggregation, quality control and splitting in Table \ref{tab:source-screening}.

\begin{table}[!ht]
\centering
\footnotesize
\setlength{\tabcolsep}{5pt}
\renewcommand{\arraystretch}{1.08}
\resizebox{\columnwidth}{!}{
\begin{tabular}{@{}lr@{}}
\hline
Item & Count \\
\hline
Original source items & 80,694 \\
After reformulatability screening & 30,669 \\
Final retained source items & 11,842 \\
Train source items & 9,476 \\
Test source items & 2,366 \\
Final item-format-paraphrase instances & 142,104 \\
\hline
\end{tabular}
}
\caption{Overall source-item and instance counts. Each retained source item is expanded into 12 final instances (4 formats $\times$ 3 paraphrases).}
\label{tab:source-screening}
\end{table}

\onecolumn
\subsection{Screening Criteria and Pipeline}

\begin{algorithm}[H] 
\caption{Multi-format data construction} 
\label{alg:multi_format}
\begin{algorithmic}[1] 

\renewcommand{\algorithmicrequire}{\textbf{Input:}}
\renewcommand{\algorithmicensure}{\textbf{Output:}}

\Require source item $x$
\Ensure retained rewritten variants or \textsc{Drop}

\Statex 

\If{$\text{ApproveCount}(x) < 3$}
    \State \Return \textsc{Drop}
\EndIf

\For{\textbf{each} $f \in \{\text{MCQ}, \text{TF}, \text{FIB}, \text{OPEN}\}$}
    \For{$k \gets 1$ \textbf{to} $3$}
        \State $\text{saved} \gets \textsc{False}$
        \State $\text{retry} \gets 0$
        
        \While{$\text{retry} \le 2$ \textbf{and} $\neg\text{saved}$}
            \State $c \gets \text{Rewrite}(x, f)$
            \State $s_1 \gets \text{Review}(c, \text{reviewer}_1)$
            \State $s_2 \gets \text{Review}(c, \text{reviewer}_2)$
            
            \If{$s_1 = \textsc{Pass}$ \textbf{and} $s_2 = \textsc{Pass}$}
                \State $\text{Save}(c)$
                \State $\text{saved} \gets \textsc{True}$
            \ElsIf{$s_1 \neq s_2$}
                \State $s_a \gets \text{Arbitrate}(c, s_1, s_2)$
                \If{$s_a = \textsc{Pass}$}
                    \State $\text{Save}(c)$
                    \State $\text{saved} \gets \textsc{True}$
                \EndIf
            \EndIf
            \State $\text{retry} \gets \text{retry} + 1$
        \EndWhile
        
        \If{$\neg\text{saved}$}
            \State \Return \textsc{Drop}
        \EndIf
    \EndFor
\EndFor

\State \Return \textsc{Retain}

\end{algorithmic}
\end{algorithm}

\onecolumn
\subsection{Prompt Templates}
\label{app:prompt-templates}

This section provides the prompt templates used in our rewriting and validation pipeline. We first show the prompt for rewriteability screening, followed by prompts for generating different answer formats. We then present the prompts used for format-specific review and final arbitration.

\begin{figure}[!ht]
\centering
\includegraphics[width=0.98\linewidth,height=\textheight,keepaspectratio]{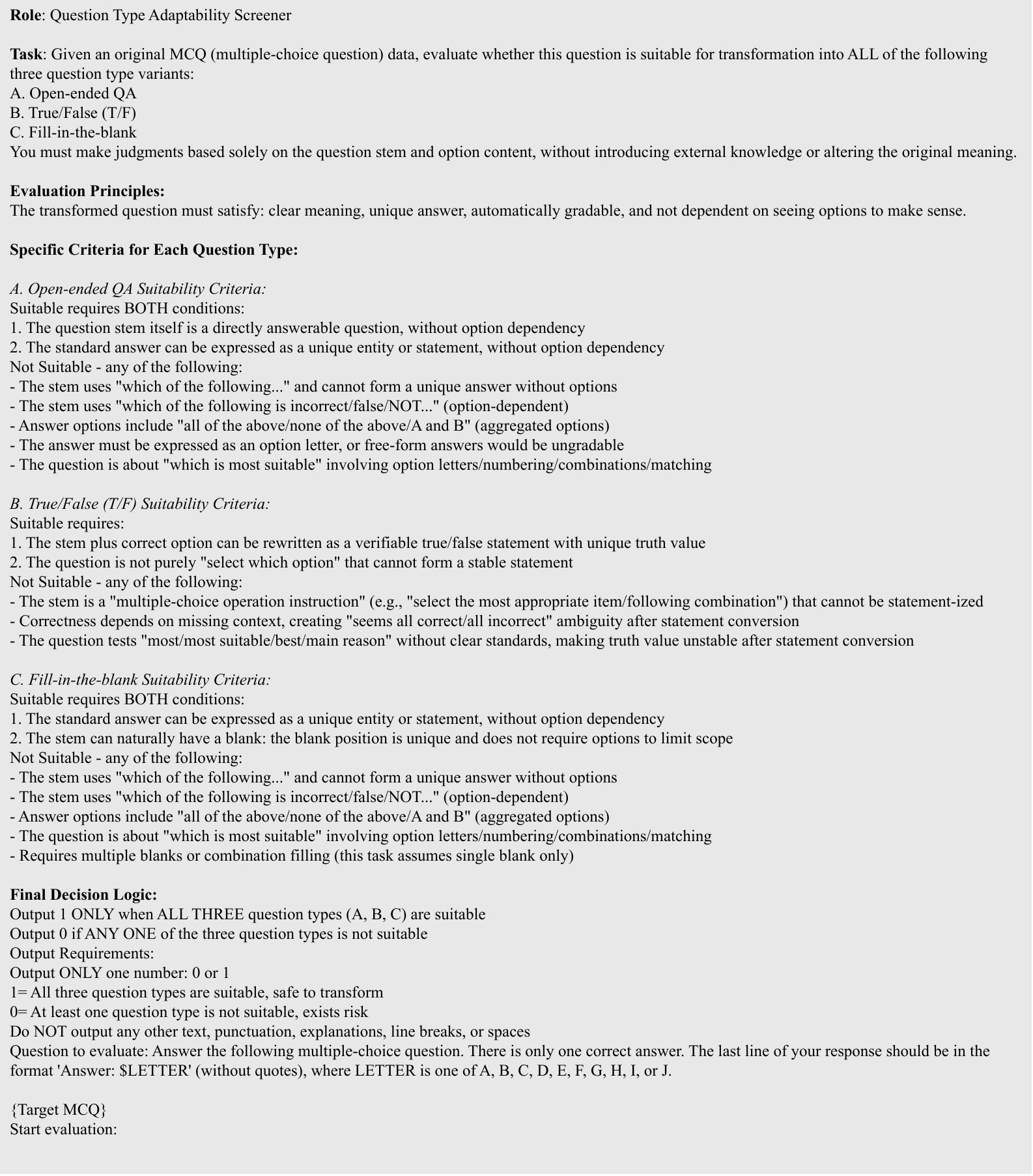}
\caption{Prompt used for rewriteability screening.}
\label{fig:appendix-filtering-prompt}
\end{figure}

\begin{figure}[!htbp]
\centering
\includegraphics[width=0.98\linewidth,height=\textheight,keepaspectratio]{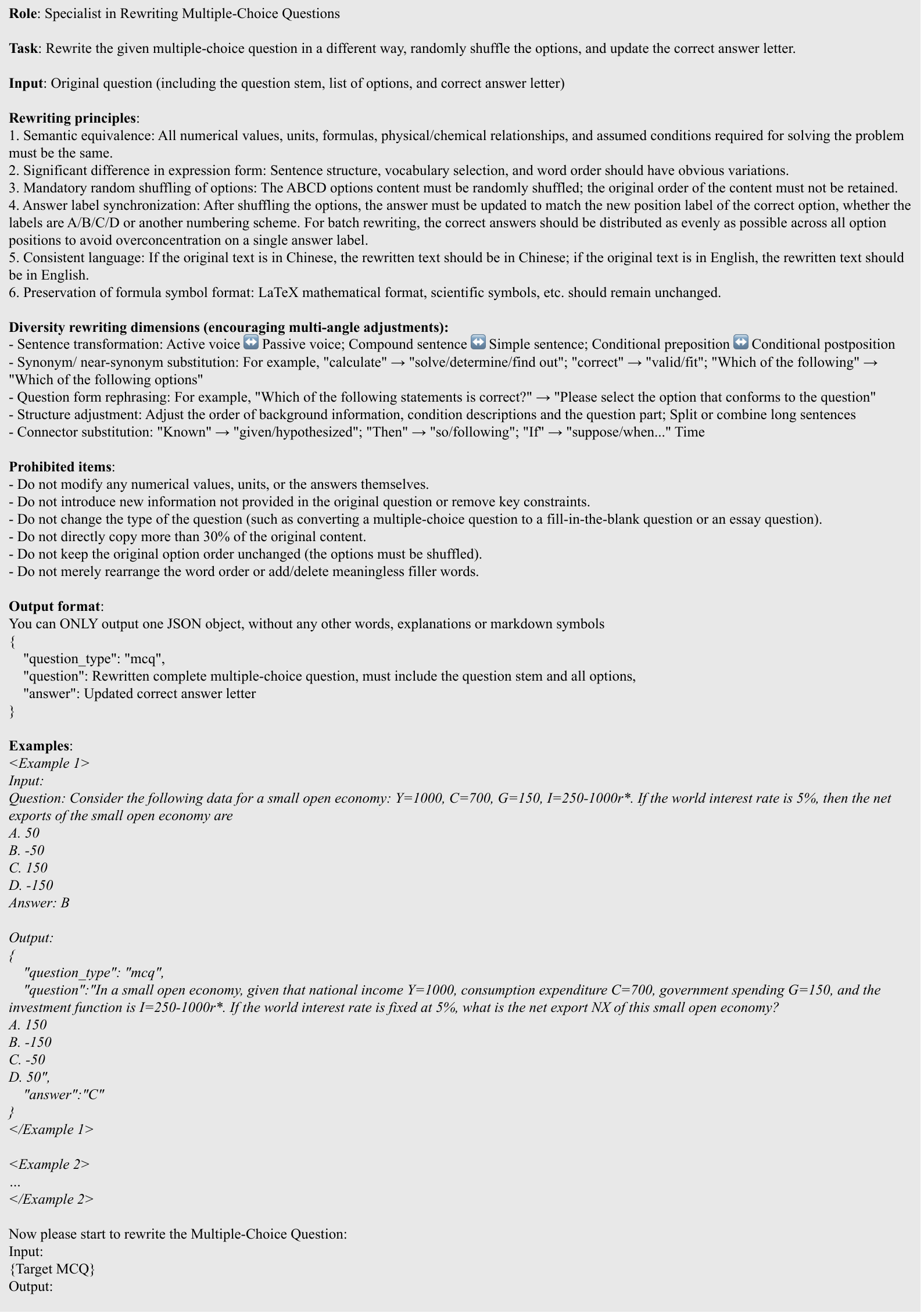}
\caption{Prompt used to rewrite MCQ items.}
\label{fig:appendix-mcq-prompt}
\end{figure}

\begin{figure}[!htbp]
\centering
\includegraphics[width=0.98\linewidth,height=\textheight,keepaspectratio]{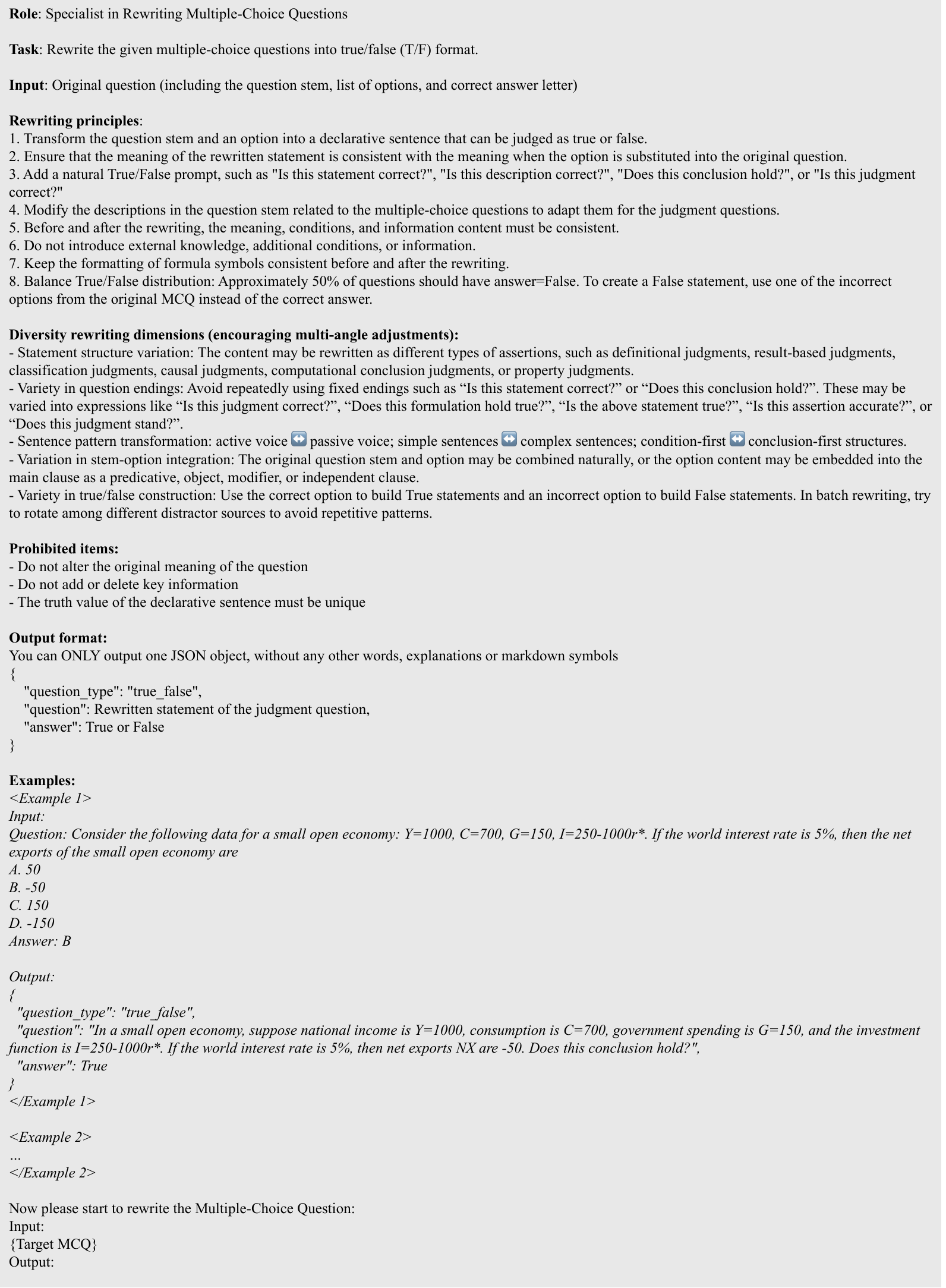}
\caption{Prompt used to rewrite MCQ items into true/false format.}
\label{fig:appendix-tf-prompt}
\end{figure}

\begin{figure}[!htbp]
\centering
\includegraphics[width=0.98\linewidth,height=\textheight,keepaspectratio]{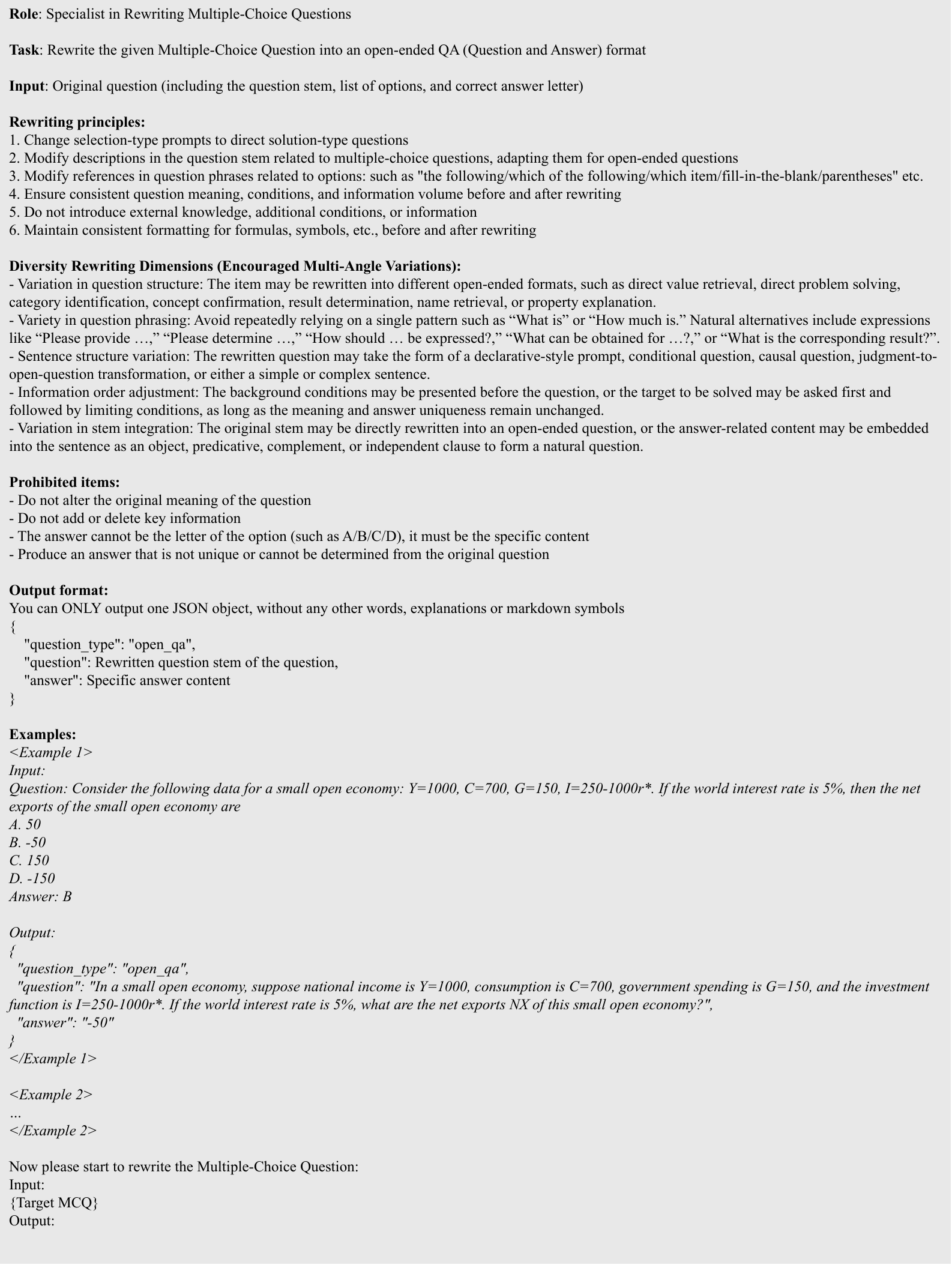}
\caption{Prompt used to rewrite MCQ items into an open-ended format.}
\label{fig:appendix-open-prompt}
\end{figure}

\begin{figure}[!htbp]
\centering
\includegraphics[width=0.98\linewidth,height=\textheight,keepaspectratio]{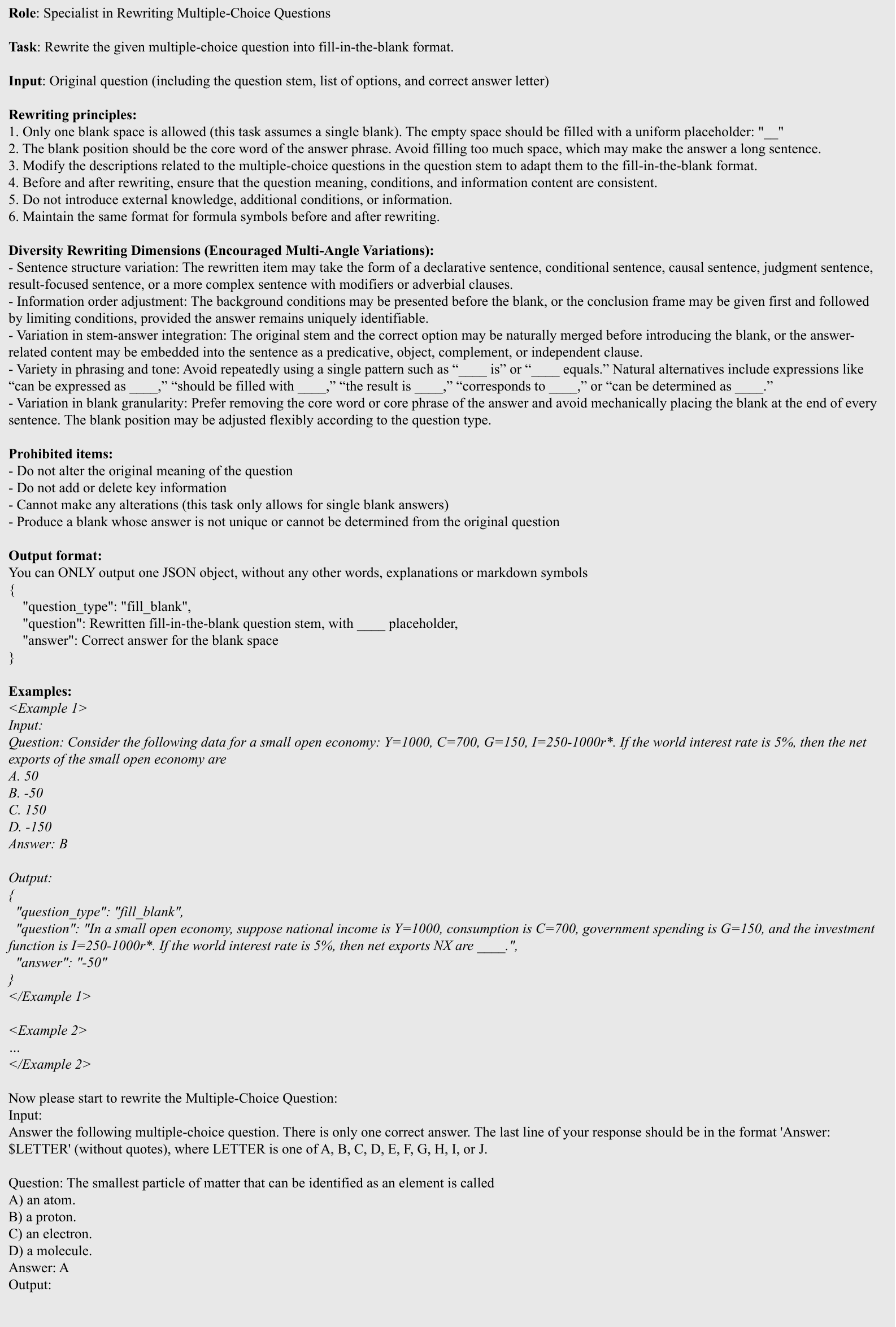}
\caption{Prompt used to rewrite MCQ items into fill-in-the-blank format.}
\label{fig:appendix-fib-prompt}
\end{figure}

\begin{figure}[!htbp]
\centering
\includegraphics[width=0.98\linewidth,height=0.9\textheight,keepaspectratio]{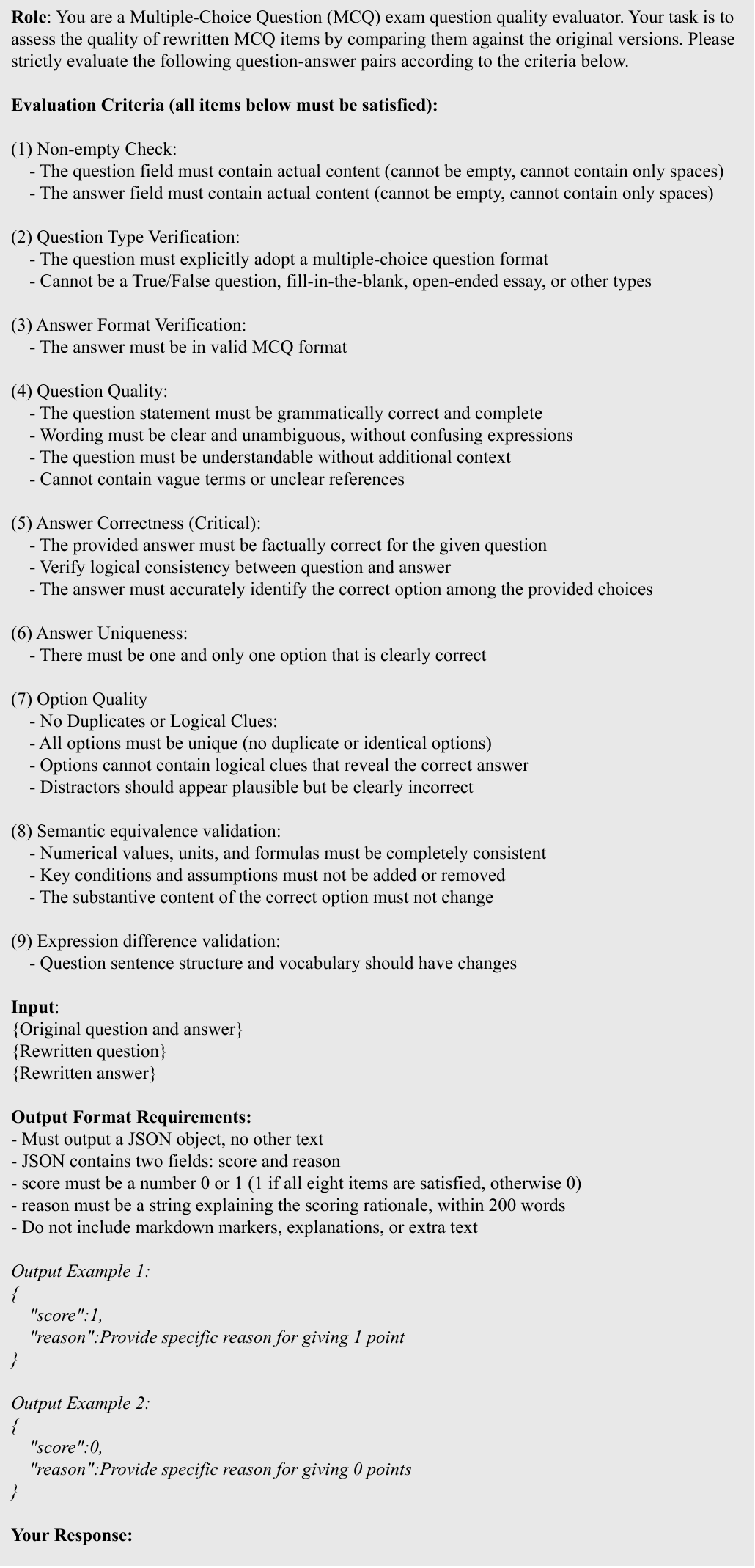}
\caption{Prompt used to review MCQ rewrites.}
\label{fig:appendix-reviewer-mcq-prompt}
\end{figure}

\begin{figure}[!htbp]
\centering
\includegraphics[width=0.98\linewidth,height=0.9\textheight,keepaspectratio]{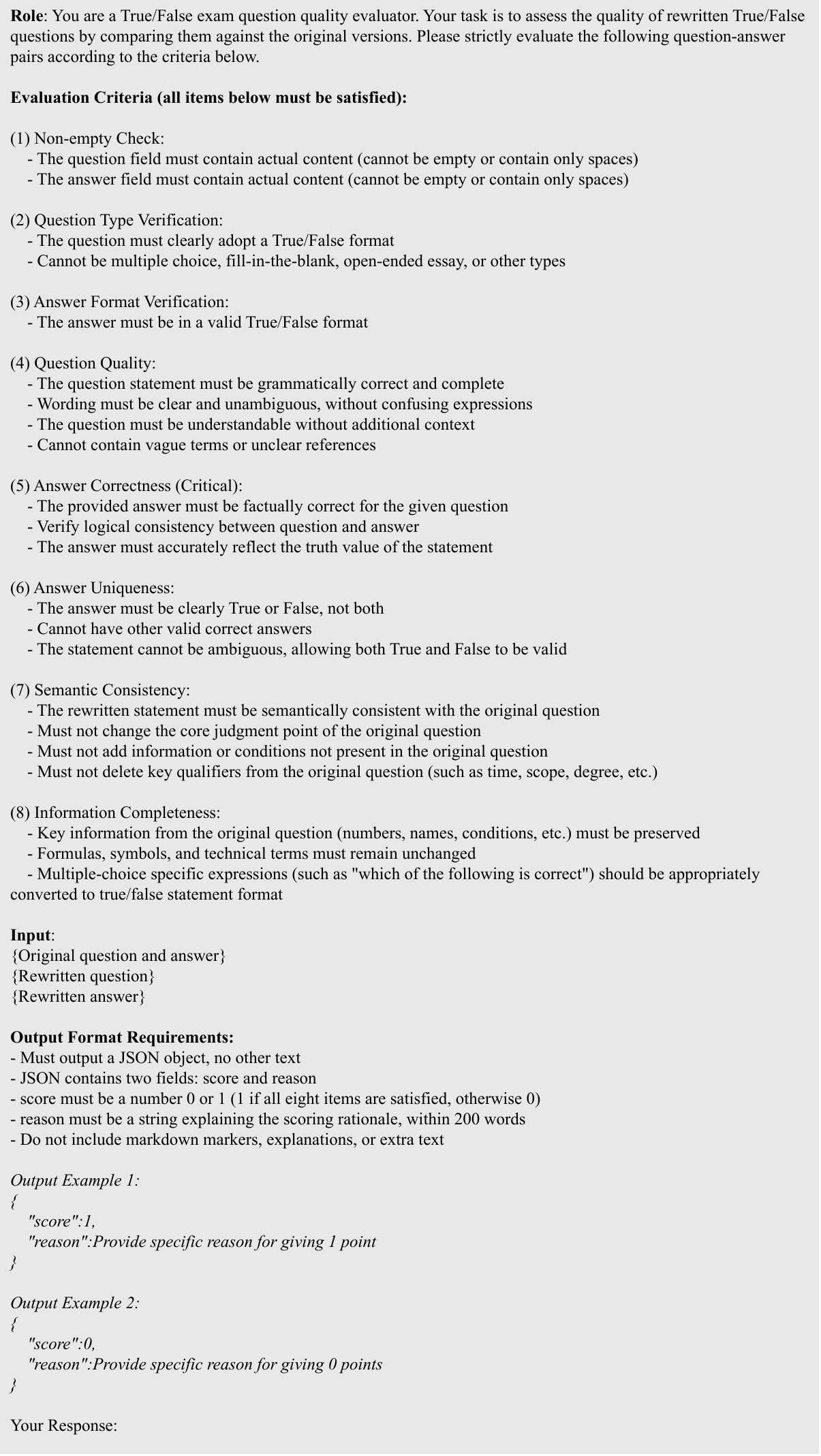}
\caption{Prompt used to review true/false rewrites.}
\label{fig:appendix-reviewer-tf-prompt}
\end{figure}

\begin{figure}[!htbp]
\centering
\includegraphics[width=0.98\linewidth,height=0.9\textheight,keepaspectratio]{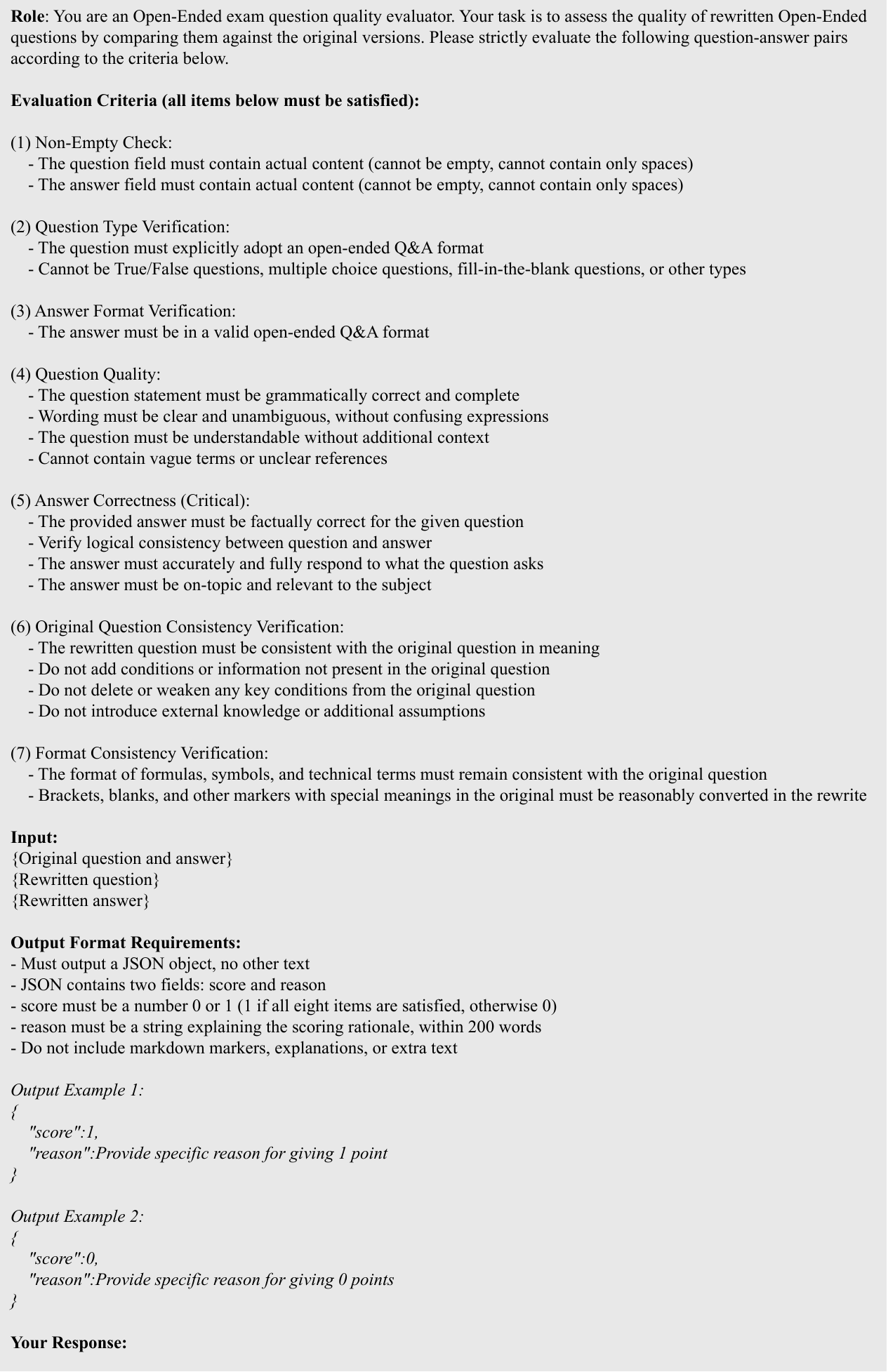}
\caption{Prompt used to review open-ended rewrites.}
\label{fig:appendix-reviewer-open-prompt}
\end{figure}

\begin{figure}[!htbp]
\centering
\includegraphics[width=0.98\linewidth,height=0.9\textheight,keepaspectratio]{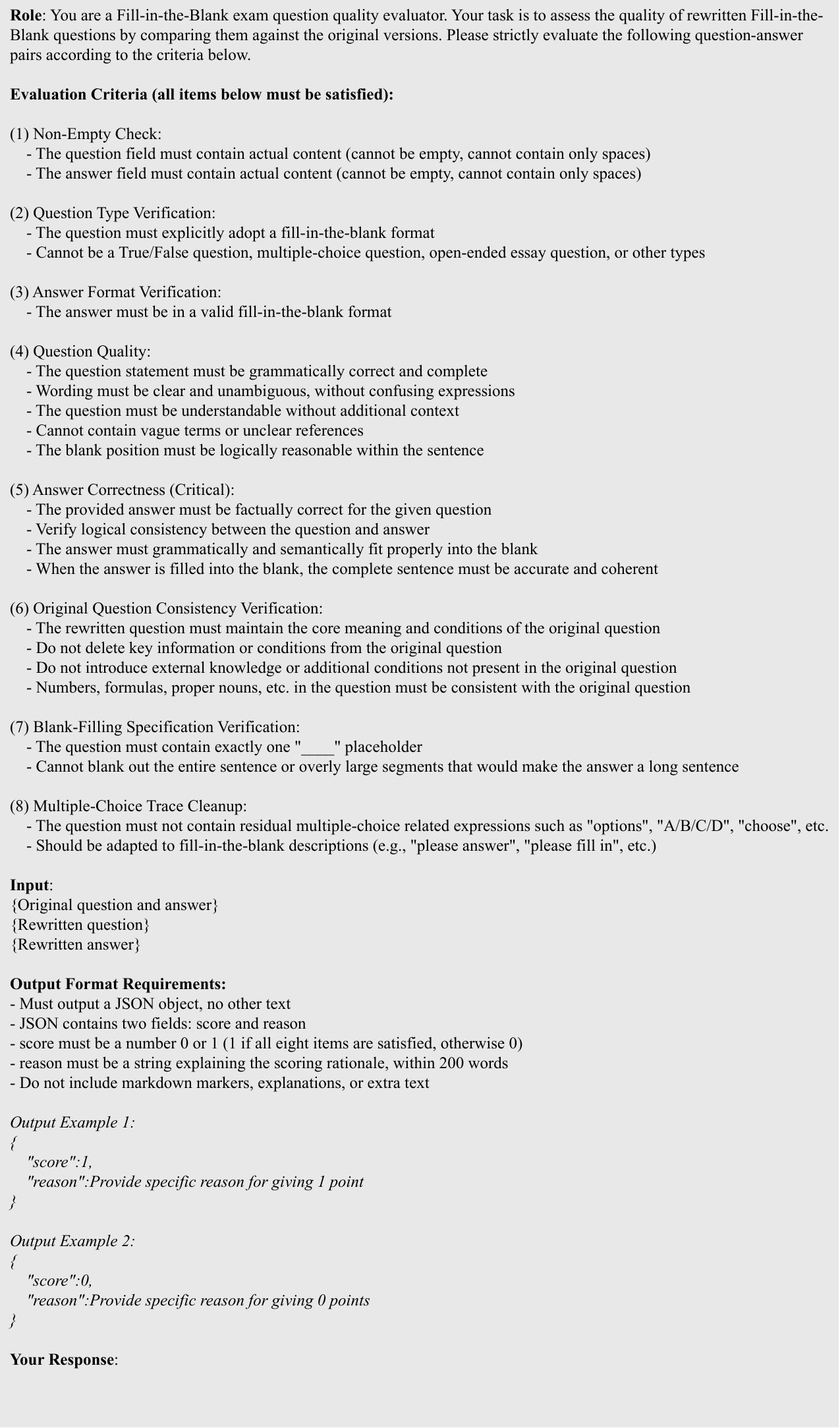}
\caption{Prompt used to review fill-in-the-blank rewrites.}
\label{fig:appendix-reviewer-fib-prompt}
\end{figure}

\begin{figure}[!htbp]
\centering
\includegraphics[width=0.98\linewidth,height=0.78\textheight,keepaspectratio]{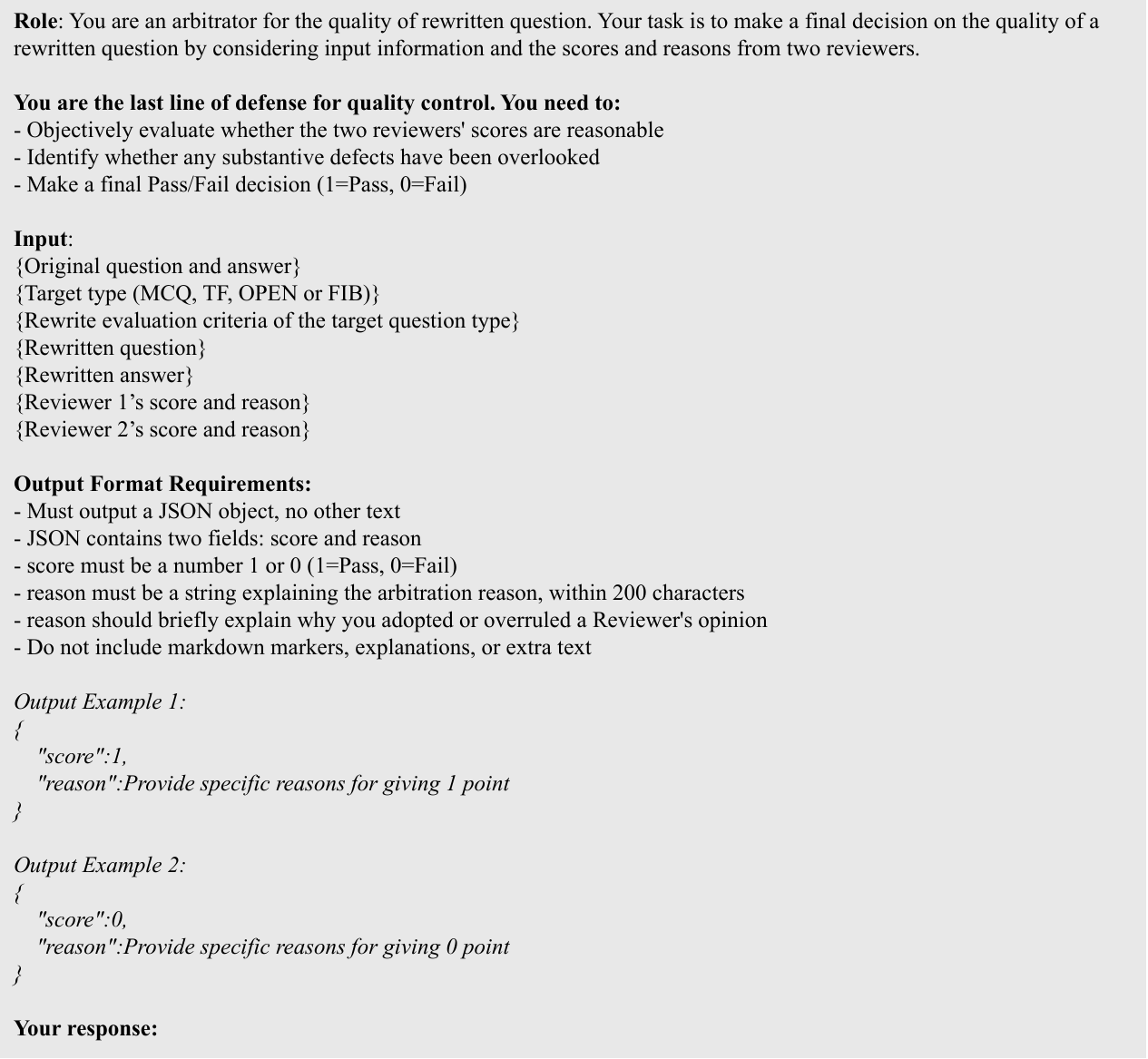}
\caption{Prompt used for final arbitration of rewrite quality.}
\label{fig:appendix-arbitrator-prompt}
\end{figure}

\clearpage